\documentclass[10pt, conference]{IEEEtran}  

\IEEEoverridecommandlockouts                              

\usepackage{graphicx}
\usepackage{caption}
\usepackage{hyperref} \usepackage{url}      
\usepackage{amsmath}
\usepackage{xcolor,soul}
\usepackage{soul}

\title{\LARGE \bf
Operating data of a specific Aquatic Center as a Benchmark for dynamic model learning: search for a valid prediction model over an 8-hour horizon
}

\author{François Gauthier-Clerc$^{1, 2, 3}$, Hoel Le Capitaine$^{2, 4}$, Fabien Claveau$^{2, 3}$, Philippe Chevrel$^{2, 3}$
\thanks{$^{1}$Purecontrol, 68 Av. Sergent Maginot, 35000 Rennes, FRANCE
        {\tt\small \{firstname\}.\{surname\}@purecontrol.com}}%
\thanks{$^{2}$LS2N – Laboratory of digital sciences of Nantes, UMR CNRS 6004, BP 92208, 44322 Nantes}
\thanks{$^{3}$IMT Atlantique, CS 20722, 44307 Nantes, FRANCE
        {\tt\small \{firstname\}.\{surname\}@imt-atlantique.fr}}%
\thanks{$^{4}$Nantes Université, Polytech’Nantes, 44306 Nantes
        {\tt\small \{firstname\}.\{surname\}@univ-nantes.fr}}%
}

\IEEEoverridecommandlockouts
\IEEEpubid{\makebox[\columnwidth]{978-1-5386-5541-2/18/\$31.00~\copyright2018 IEEE \hfill} \hspace{\columnsep}\makebox[\columnwidth]{ }}

\begin{document}

\maketitle
\IEEEpubidadjcol


\begin{abstract}

This article presents an identification benchmark based on data from a public swimming pool in operation. Such a system is both a complex process and easily understandable by all with regard to the stakes. Ultimately, the objective is to reduce the energy bill while maintaining the level of quality of service. This objective is general in scope and is not limited to public swimming pools. This can be done effectively through what is known as economic predictive control. This type of advanced control is based on a process model. It is the aim of this article and the considered benchmark to show that such a dynamic model can be obtained from operating data. For this, operational data is formatted and shared, and model quality indicators are proposed. On this basis, the first identification results illustrate the results obtained by a linear multivariable model on the one hand, and by a neural dynamic model on the other hand. The benchmark calls for other proposals and results from control and data scientists for comparison.

\end{abstract}


\section{Introduction}
The need for energy and commodities is constantly increasing and must be carefully managed to avoid possible future shortages due to limited resources. From an energetic perspective, we observe a switch from highly controllable electric plants powered by fossil fuel to renewable non-controllable sources like wind and solar. This paradigm shift will introduce more discrepancies between demand and production, leading to more volatile prices. From a physical/hardware aspect, improving existing plants, or building new modern efficient plants, become more and more expensive in the context of commodities inflation. Despite the fact that they lead to a lot of savings, these methods let existing systems aside.

These observations show the need to improve actual plant efficiency by retrofitting process control to allow more consumption flexibility and global efficiency. This approach does not involve a lot of investment since we are dealing with software upgrades only. 

Today, a lot of energy-intensive plants are controlled by a simple controller without any economic optimality concern \cite{DRGONA2020190}. 
Although numerous optimal controllers are available in the academic world \cite{mpc_piscine_eff,ALAJMI2012122}, their deployments are still expensive and not well-established in some parts of the industrial world. This situation is even more pronounced in small and medium facilities like swimming pools or small HVAC (Heating Ventilation, and Air Conditioning) systems due to a lack of resources allocated to control. 

The challenge is therefore to propose advanced control methods that are efficient, have low development/deployment costs and are suitable for a wide variety of industrial systems, taking advantage of their common characteristics. The temperature management of swimming pools is part of this scope.

Today, many regulation systems (especially with swimming pools) are based on a closed-loop PID system with a fixed working point. It is a reliable and cheap regulation, which explains its popularity.  For the swimming pool case, we observe a constant 28-degree target most of the time \cite{instance1290}. Although this strategy is popular, its lack of economical optimality tends to question its use in the context of energy restrictions (implying an increasing cost) and non-permanent temperature demand (closed at night). More advanced expert based-technique are available to achieve better regulation. They rely on scheduling methods to reduce the average temperature in order to make savings (decreasing the average temperature can save up to 7\% energy demand per degree \cite{DelgadoMarn2020DynamicSM}). They are based on static \cite{ESCRIVAESCRIVA20102258} or dynamics temperature \cite{lee2008data,Zemtsov2017} scheduling methods to update the setpoint according to demand.

Unfortunately, their approaches are still simple and mainly rely on expert knowledge to formulate strategies. The validity of open loop schedulings is affected by exogenous effects like outdoor temperature or other disturbances \cite{henze2005predictive} which make them expensive to deploy and maintain. This situation highlights the need for other techniques, like model-based control. Instead of using expert knowledge for the control design, we can use a data-based system identification method \cite{ALARAJ2022103678} that allows model-based control methods. These methods generate a model-based control directly on the constraints (humanly interpretable). This approach avoids the need for experts and therefore reduces the deployment cost for equivalent accuracy in some tasks.

In the context of system control with economical awareness, the EMPC (Economical Model Predictive Control) algorithm has gained more and more interest in recent years. As explained earlier, this control method does not focus on setpoint (like tracking MPC) but uses an economical function instead. There are several contributions to HVAC systems \cite{DRGONA2020190} and swimming pool \cite{mpc_piscine_eff}.
For medium and small facilities like swimming pool, this control method is simple to do (small state space) and allow economical saving \cite{Caspari2020}.

In the field of swimming pools, there is a lack of study that involves data-driven approaches to control. However since some interactions, like evaporation, are difficult to model \cite{SHAH2012306}, data-based methods are even more justified. Some work seems to show this interest in the context of public swimming pools \cite{LOVELL2019106167}.

In order to help the academic community to develop new frugal energy management solutions, this article defines the outline of a benchmark,  by making available the operating data of a public swimming pool. It makes it possible to consider the problem of learning dynamic models for control, MPC in particular. The proposed dataset is interesting for several reasons. It comes from the sensors which are conventionally fitted to this type of installation, in operation, over a long period.
This dataset also contains different seasons which have an important impact on the energy balance of the installation and add a realistic complexity.
To initiate the benchmark, we propose two identification results as a baseline for future contributions.

This document is organized as follows; Section 2 introduces the swimming pool system with a description of the use case and some overviews of the physical models, for a better understanding of the process. Section 3 details the data set and assessment criteria based on the need for data-driven control. Section 4 presents the primary results obtained for two selected black-box models.

\section{The public swimming pool system: Water temperature regulation}

\subsection{Benchmark case study}

\begin{figure*}
  \center
  \includegraphics[width=0.8\textwidth]{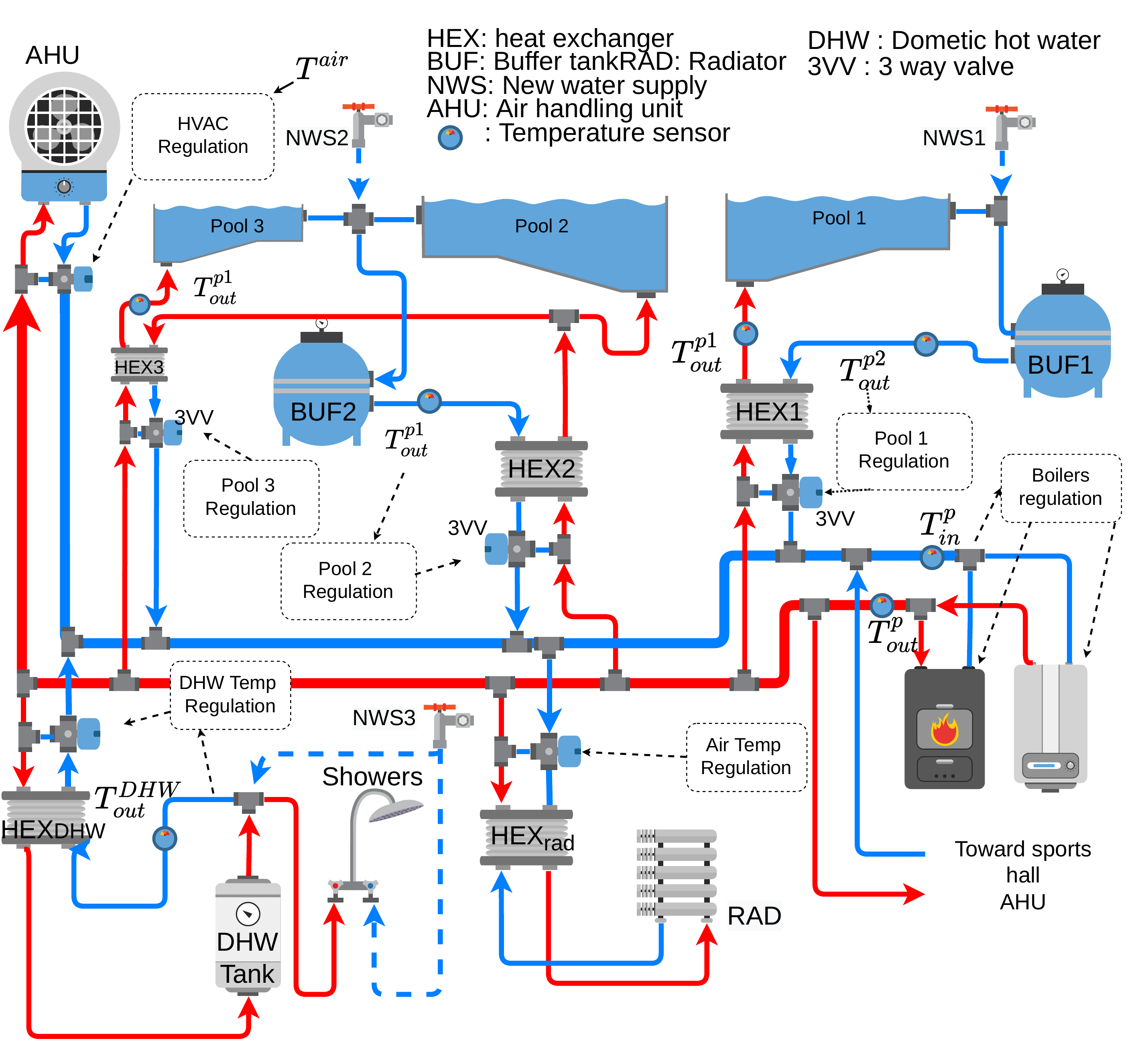}
  \caption{Diagram of the heating network}
  \small{The diagram represents the water system only. The water pumps are not shown due to lack of space. In the real system, there are several radiator networks as well as small auxiliary pools using the same system as pool 1.}
  \label{fig:schema_full}
     \vspace{-5pt}

\end{figure*}

A swimming pool is a plant that needs to regulate pool temperature and quality as well as air humidity and temperature. It uses several local regulations to maintain water quality, air state and water pool temperature (see Fig. \ref{fig:schema_full}). For the air and pool temperatures, the plant uses a heating system that consists of a primary hot water loop to distribute energy across the plant from a given source of energy (mostly a dedicated boiler). Boilers are regulated thanks to some local controllers. In practice, each regulation is done with a local hardware regulator provided by each sub-part manufacturer. Global regulations for this type of facility are very rare or even non-existent.

The water constantly circulates from each pool to its recycling system in order to: heat, clean and store the water. The water quality is maintained by UV lamps and filters. The system also includes a buffer tank to store the water when people are using the pool (less water volume is needed). The water is also renewed with new clean water. The fresh water flow rate is based on the measured filter pressure and is activated 2 or 3 times a day for a short period of time. This water refilling is the main perturbation of this system, as the fresh water is very cold compared to the pool temperature. The water is heated by a countercurrent heat exchanger and the temperature is regulated according to temperature sensors before the heat exchanger. The temperature regulation is based on a 3-way valve that bypasses or not the water flow rate in the hot side of the exchanger. The water flow rate in the recycling loop is maintained with a pump and overflow mechanism. The flow rate setpoint is often constant but can change according to static scheduling (day/night). This flow rate can affect the dynamic of the heating process.

The aimed facility is a public swimming pool located in Brittany, FRANCE. This swimming pool is composed of two main pools (sport and recreational) as well as one hot tub and paddling pool. The sports pool has a volume of 672 m$^3$ (12.5m $\times$ 25m) with 5 lanes. This pool is coupled with a buffer tank of 30m$^3$. The second pool (of 212 $m^3$) is coupled with the paddling pool (of 12$m^3$). They share the same recycling flow and a buffer tank of 35 m$^3$. 

\begin{table}[ht]
    \centering
    \begin{tabular}{lcc}
      \hline
        Characteristic &  Value &  Unit \\
              \hline
        Pool 1 volume & 672 & $m^3$\\
        Pool 1 depth & 1.3 to 3.0 & m\\
        Pool 1 recycling flow rate & 189 & $m^3/h$\\
        Pool 1 buffer tank size & 30 & $m^3$ \\
        Pool 2 volume & 212 & $m^3$\\
        Pool 2 depth & 0.45 to 1.3 & m\\
        Pool 2 recycling flow rate & 142 & $m^3/h$ \\
        Pool 2 buffer tank size & 35 & $m^3$ \\
        \hline
    \end{tabular}
    \caption{Swimming pool characteristics}
    \label{tab:caract_system}
\end{table}

The swimming pool is open during the whole year from 7 a.m to 5:30 p.m each day. We observed between 250 and 600 swimmers across the day. The temperature requirement is constant during the opening period, 28 degrees for the sports pool and 31 for the recreational one. On certain Wednesdays throughout the year, baby swimming lessons are organized and require a high temperature of 32 degrees in the second pool. All specifications are available in the TABLE \ref{tab:caract_system}.


In this facility, the heating system consists of a wood boiler and a gas boiler. The first one is used as a priority, and the second one is used when there is a shortage of wood or a high energy demand. The wood boiler provides a max power of 650kW and the gas boiler a power of 350kW. A peak energy production (with both boilers) can reach around 1MW.

In addition to the double boiler system, a sports hall uses the energy network for its air heating system. This particular energy consumption is monitored with dedicated sensors.

This public swimming pool gathers all the particularities of such a system (coupled pool, dual boiler, external energy consumption).  It is complex enough to require special attention if traditional expert approaches are used to design the supervision system, regardless of the methodology considered; simulation-based approach or expert control rules. One can expect data-driven approaches to be able to handle all the details in an automatic process.

\subsection{Identification needs for local optimal regulation}

Based on our earlier motivation, the easiest and cheapest way to improve these simple regulations is to improve only one or two regulators. This approach reduces the deployment cost and can save a large part of the possible gain. From this idea, it is obvious to take control of the pool temperature regulation in order to save energy. We choose to take care of the two larger pools of the facility; Pool 1 (sports pool) and Pool 2. The Pool 3, the paddling pool regulation is ignored since it represents only 5\% of the second pool volume.

The pool temperature can be manipulated with the control variable (3-way valve) in order to reduce the average temperature (reduce energy losses) or store energy when it is cheap while staying compliant with the user's requirements (thanks to predictive control).

The main action to save energy is to reduce the temperature when the pool is closed. In order to do this optimally, it is important to reheat the temperature early enough to achieve a compliant temperature at the pool opening. In this context, it is crucial to have a model that can predict the right time to heat or choose the pool with the highest energy-saving potential (if we have a limited boiler power). In this context, an 8-hour prediction according to the potential heating time is chosen for this benchmark. This horizon is large enough to handle the morning heating which can be a slow process.
The model also needs to take into account exogenous data (such as outside temperature, air moisture, etc.) to be reliable over different periods. 
Moreover, the open loop 3-way valve and temperature are chosen to allow more efficient control (be able to saturate the control for instance). For an illustration, please refer to \cite{Caspari2020}.

To be able to manage all pools, the model needs to take into account the set of pools as a whole. Indeed, boilers have limited power which means that pools can interact with each other through energy consumption. This situation leads us to a MIMO formulation in order to deal with their problems and remain in a purely black box approach. 

Before presenting the dataset, let us describe the theoretical model of pool energy balance which helps to understand macroscopic physical phenomena and data selection. This theoretical model, although enlightening, is insufficient. It will not be used in the process of identifying the system, the hypothesis retained being to proceed from the data; without having recourse to physical knowledge of the process to be controlled, and thus keep the cost of deployment cheap.

\subsection{Theoretical aspect}

The pool dynamics can be summarized by its heat balance, taking into account all sources of heat loss and gain.

\begin{equation}
V_{pool}  \rho_w c_w \frac{dT_{pool}}{dt} = \dot{Q}_{tot} 
\end{equation}

In a traditional indoor swimming pool, we can  sum up all of these interactions as follows:
\begin{equation}
\dot{Q}_{tot}   = \dot{Q}_{evap} + \dot{Q}_{rad} +  \dot{Q}_{cond} + \dot{Q}_{conv} + \dot{Q}_{refill} + \dot{Q}_{HE} 
\end{equation}

From the equation above, we have evaporation ($\dot{Q}_{evap}$), radiation ($\dot{Q}_{rad}$), conduction ($\dot{Q}_{cond}$), convection ($\dot{Q}_{conv}$), water recycling (or refill) ($\dot{Q}_{refill}$) and heat exchanger ($\dot{Q}_{HE}$) contribution to the temperature dynamic ( see \cite{Calise2018} for more details).
Based on these interactions, the evaporation phenomenon is highly dependent on the activity of the pool, which is very difficult to measure \cite{Asdrubali2009}. The water recycling and the heat exchanger contribution are driven by the local controller.  Like water evaporation, the impact of fresh water depends on its temperature, which is rarely measured in practice and is often seasonally correlated.

The heat exchanger allows for heat transfer from the primary loop to the pool. This 3-way valve is controlled with an opening setpoint that induces a non-linear static relationship with the water flow rate \cite{Zajic2011}.  The hot water flow rate gives energy to the pool according to the recycling flow rate. This interaction can be approximated using the NTU method \cite{incropera1996fundamentals}. Energy transfer can be measured using upstream and downstream sensors, although depending on the system, the upstream sensor may not always be available (since conventional control does not require it).
The primary water circuit that carries the heat is also connected to other subsystems such as HVAC, water tanks (for showers), and radiators (see Fig. \ref{fig:schema_full}). This situation can lead to a heat shortage at some point if all of those heat consumers are calling for energy. The circuit is made in parallel in order to avoid prioritizing certain subsystems.

The air state has its own energy and humidity balance. This dynamic includes the pool hall regulation system, pool state, and heat loss to the outside. This temperature is regulated with a local controller to maintain a constant setpoint.

\begin{figure}
    \centering
    \vspace{8pt}
    \includegraphics[width=0.48\textwidth]{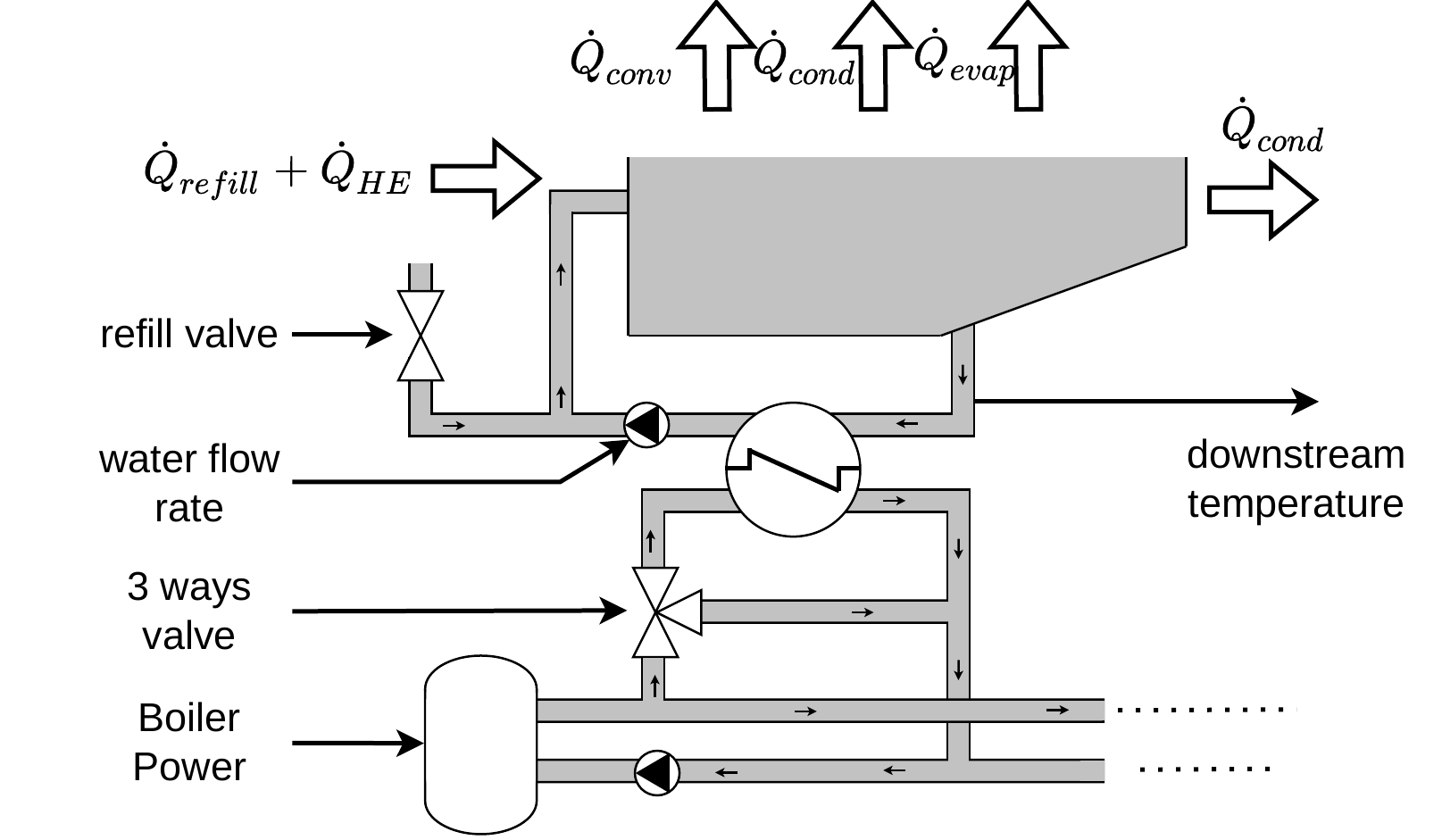}
    \caption{the pool sub system}
    \label{fig:system_schema}
    \vspace{-22pt}
\end{figure}

We summarize all of the elements in Fig. \ref{fig:system_schema}, where the input signals are on the left side and the output signals are on the right.

\section{Benchmark dataset}

\subsection{Data selection}

\begin{figure}
    \centering
    \vspace{15pt}
    \includegraphics[width=0.49\textwidth]{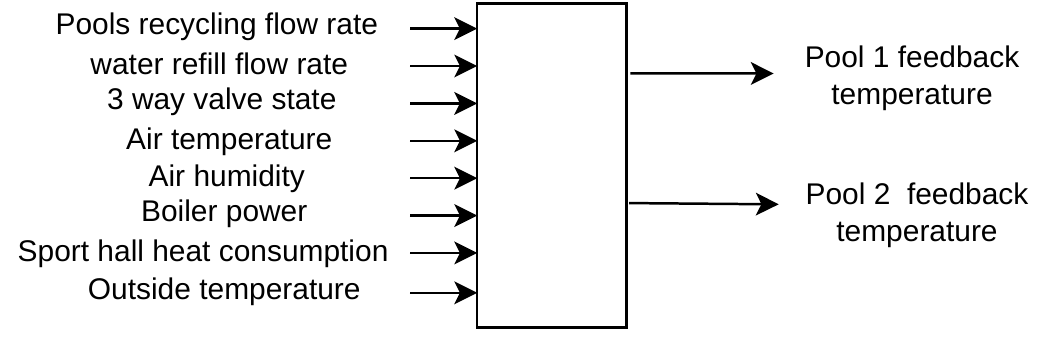}
   \caption{MIMO formulation}
   \label{fig:aaa}
      \vspace{-22pt}
\end{figure}

The main objective is to predict pool temperature using only available sensors to minimise deployment costs. We aim to use as few exogenous signals as possible to ensure that the identification can be deployed to a large number of installations.
In this context, we include the total boiler power, 3-way valve state, indoor air temperature and humidity, outdoor air temperature, recycling speed, refill water flow rate, and outside temperature as input signals. The total boiler output is measured using a dedicated sensor that can encapsulate the complexity of the heating system. If this sensor is not available, it can be deduced from the input/output temperature and the flow rate of the primary loop. The air temperature and humidity are relevant for heat losses with air. The 3-way valve is the control signal, and the new water flow rate is a perturbation. Despite all this information, some data are not easy to capture and are not included in this study. For example, wind speed \cite{mpc_piscine_eff} is necessary for evaporation modeling, as well as pool attendance, but they are not available here. We also include the sports hall energy consumption in this study since it can affect the heating power for the pools. This variable is the only one that is specific to this installation. The sensor selection is summarized in Fig. \ref{fig:aaa}.

The temperature of the primary loop is not included as it is part of the system state and must be used as an output signal. Including it would increase the complexity of the identification task and the integration into predictive control algorithms. This choice respects the purely black box data-driven approach and forces the identification task to include the energy transport mechanism in an implicit manner. On the other hand, the air state is more stable and can be predicted with an ad-hoc algorithm (like time series prediction \cite{MENDOZASERRANO20141301}).

\begin{figure*}
    \centering
    \vspace{8pt}
    \includegraphics[width=0.9\textwidth]{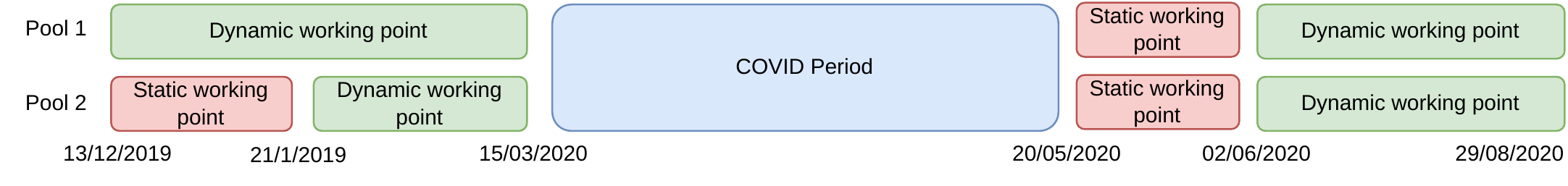}
        \vspace{-8pt}

    \caption{Dataset timeline}
    \label{fig:timeline}
    \vspace{-15pt}
\end{figure*}

The data is subsampled using a moving average, and the signal is processed to remove sensor faults. Due to missing data and special events like COVID-19 lockdown (see section \ref{sec:context}), the dataset is split. The data is also normalized according to common practice \cite{ljung1998system}.  The timeline in Fig. \ref{fig:timeline} summarises all this information as well as the type of control. 

The dataset includes both seasons, winter and summer. The temperature setpoint for pool 1 fluctuates between 27 and 28 degrees, while for pool 2, it fluctuates between 30 and 31 degrees.

\subsection{Context}
\label{sec:context}

\begin{figure}
  \center
  \includegraphics[width=1.0\linewidth]{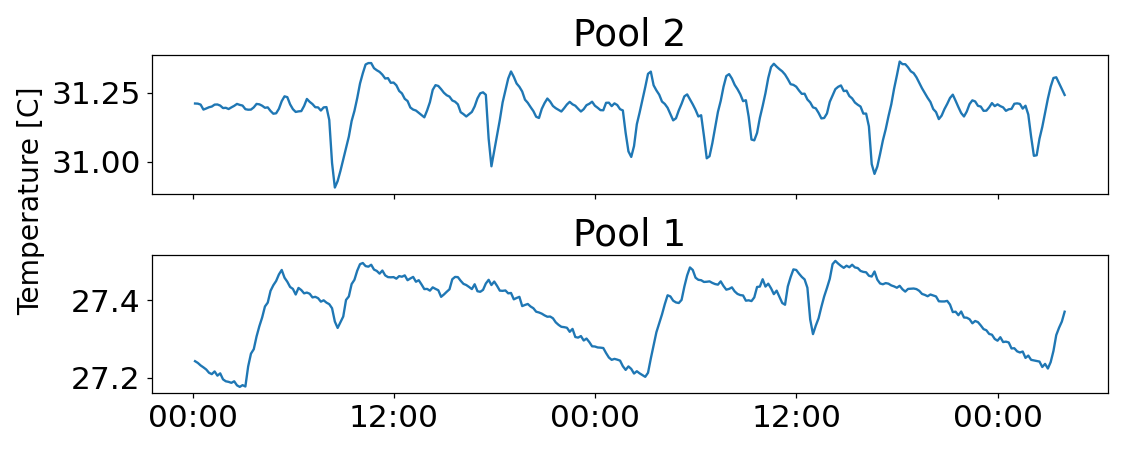}
  \caption{Illustration of pool regulation of the dataset}
  \label{fig:illustration_data}
  \vspace{-5pt}
\end{figure}

The data was collected using several hardware automatons, with each signal being measured every minute and stored since 2018.  The benchmark uses data from September 2019 to August 2020. We keep this range since the swimming pool has been extended with an outside pool that complicates the analysis (more exogenous data to take into account). Moreover, this period contains several special events that generated good data diversity for model validation. Among these events, the COVID-19 period forced the pool to close for 3 months, which created a period with a low temperature. In addition, the pool uses two control strategies: one PI regulator with a constant setpoint and a more advanced strategy that reduces the temperature at night.
This second method allows for better data for identification since the control signal is not correlated with the feedback signal, and it closely approximates the optimal strategy. Both types of regulation can be compared in Fig. \ref{fig:illustration_data}. In this figure, Pool 2 is regulated according to a static setpoint, while Pool 1 is regulated with a dynamic setpoint that allows for a reduced temperature at night. The static setpoint regulation does not provide much information, as the only perturbation to the temperature is from the addition of new fresh water. In contrast, the dynamic regulation strategy adds a lot of diversity and implements the energy-saving strategy mentioned earlier.

\begin{figure}
  \center
  \includegraphics[width=1.0\linewidth]{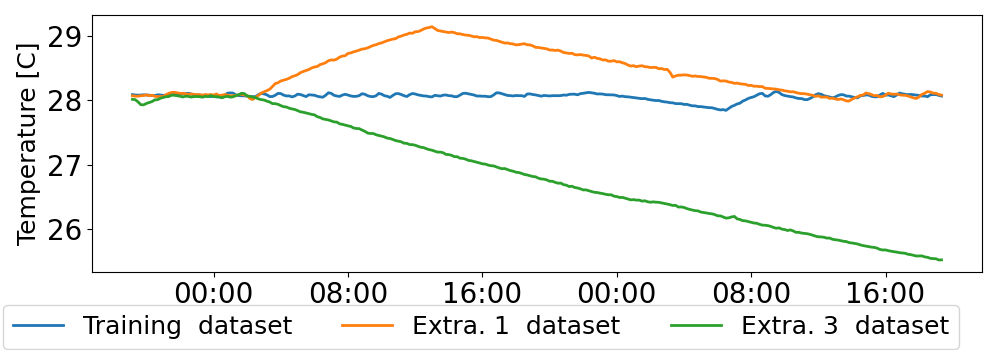}
  \caption{Illustration of two different extrapolation scenarios for Pool 1, compared to one period of the training data.}
  \label{fig:illustration_data_extra}
  \vspace{-15pt}
\end{figure}

Furthermore, this period includes four situations where the temperature was managed differently than usual, which we refer to as "scenarios". These scenarios were caused by factors such as human error and sensor faults and can be considered as potential planning scenarios for model-based implicit controllers such as EMPC. Two of these scenarios (scenario 1 and scenario 3) for the first pool are represented in Fig. \ref{fig:illustration_data_extra}. Note that these scenarios were excluded from the training dataset.

The four scenarios are as follows:
\begin{itemize}
    \item  A stuck 3-way-valve on pool 1 at 50\%.
    \item Pool 2 cooled down to 25 degrees while pool 1 remained at its nominal temperature.
    \item  Pool 1 cooled down to 25 degrees while pool 2 remained at 25 degrees.
    \item Both pool 1 and pool 2 were heated up to their nominal temperature at the same time period.
\end{itemize}


\subsection{Identification problem \& Evaluation}

The identification problem is closely linked to the use that will be made by the control (MPC) of the resulting model. The identification is therefore evaluated through a prediction horizon which is not restricted to one step, as is conventionally done. Moreover, the current state and the exogenous signals are assumed to be known.

The 10-minute sampling time used leads to a prediction through 48 steps to reach the 8-hour horizon. The prediction accuracy over the 48 steps horizon can be given with the function $\mathcal{L}(1, H)$, defined below:

\begin{equation}
    \mathcal{L}(I, J) = \frac{1}{J-I+1}\sum_{i=I}^{J} \sqrt{\sum_{k=0}^{K-1} \frac{||\hat{y}[i+k| k] - y[i+k]||^2_2}{K}}
\end{equation}

The prediction $\hat{y}[k+i| k]$ is estimated based on the observation at instant k ($y[k]$) and exogenous and control signals from the instant $k$ to $k+i$. In this benchmark, the vector $y$ contains the temperatures of the two pools.
This horizon accuracy is obtained over the entire test set in a convolutive manner.

It is useful to visualize the accuracy of each candidate model according to the prediction depth by plotting the function $x \rightarrow \mathcal{L}(x,x)$ for $x$ in $[1,H]$ (Fig. \ref{fig:rmse_model}).

From the function $\mathcal{L}$, we need to define several criteria to validate and evaluate the model for control. To do so, we have three sections of testing data placed at different system regimes during the year of data. Secondly, we have four scenarios to evaluate the model in an extrapolation context.

In order to use $\mathcal{L}$ and enrich the analysis, three criteria have been defined to evaluate the model's prediction ability for simulation and control. The first criterion includes the entire prediction window, the second is limited to short-term prediction, and the last one is for long-term prediction. The goal is to achieve good accuracy for an 8-hour horizon with good short-term and long-term accuracy (refer to TABLE \ref{tab:acc_table}).

\begin{table}
    \centering
    \vspace{8pt}
    \begin{tabular}{|l|c|}
        \hline  Name & Formula \\ \hline
        Full horizon acc &  $\mathcal{L}(1, H)$ \\
        Short term acc   &  $\mathcal{L}(1, H/4)$ \\
        Long term acc    & $\mathcal{L}(3H/4, H)$ \\
        \hline
    \end{tabular}
    \caption{Model accuracy criteria}
    \label{tab:acc_table}
       \vspace{-15pt}

\end{table}

In addition to evaluating the model's performance on the test set, analyzing the model's extrapolation capacity is also critical in this black-box identification for control. To analyze this capacity, we use the full horizon accuracy per scenario to determine which extrapolation scenario is well-predicted or not compared to the test set.

To summarize, the goal is to achieve good accuracy in the 8-hour horizon with both short-term and long-term accuracy. The model's ability to extrapolate is critical to assess its usefulness in the control scheme. Prior physical knowledge can guide the identification process, but it is important to keep in mind the initial goal of cheap deployment across different swimming pool architectures.

All resources for this benchmark are available at  \url{https://benchmark-datadriven-sysid.purecontrol.com}. These resources include data and tools for plotting and evaluating models with Python and Matlab.

\section{Identification method \& proposed models and results}

\subsection{State of art }

System identification is a vast field that includes model estimation, model reduction, experiment design and more \cite{ljung1998system}. In this paper, we focus on formulating a suitable model in the context of black box modeling. Unlike other methods, we mainly rely on real data, which is more challenging to deal with. Among these black box models, we can cite linear polynomial model (ARX, OE, ARMAX), linear state space (LSS) \cite{ljung1998system}, neural network (NN), linear parameter varying (LPV) \cite{Thilker2022} and many more. This large dictionary allows users to choose the appropriate model complexity depending on the problem.

Identification of linear models has been carried out in both simulation \cite{5759140} and real-world contexts \cite{381251} for HVAC systems using the ARX formulation. These data-driven models are used in EMPC to manage indoor temperature based on endogenous and exogenous signals, with models capable of predicting up to 24 hours ahead. In a purely data-driven approach, the authors of \cite{FERKL2010205} utilized subspace methods to identify an entire building using sensors.

With greater expressivity, we can mention data-driven LPV (Linear Parameter Varying) identification \cite{Thilker2022} using parametric methods (cubic spline) to model non-linear exogenous effects in an HVAC context.

In recent years, pure non-linear models like neural networks have gained more attention. They are considered universal approximators that allow for a wide range of applications. Many applications in the control of HVAC systems can be cited \cite{AFRAM201796}. In this paper, the authors highlight the fact that despite the computational burden due to non-linear MPC, their application is possible with modern computers for control. Neural Networks under NLARX (Non-Linear Auto Regressive eXogenous) modeling have been successfully identified with real data \cite{AFRAM201796, Mechaqrane2004}. Neural network models often outperform linear models in terms of pure accuracy metrics \cite{AFRAM201796, Mechaqrane2004}. However, it is important to note that the generalizability property of these models with different data distributions is not guaranteed. 

As a first identification application on this proposed benchmark, it seemed to us interesting to select both a pure linear state-space model with the subspace method and a neural network-based NLARX model.

There is also a large number of hybrid models in the grey box paradigm capable of achieving good results through the addition of knowledge \cite{Zajic2011,Thilker2022}. However, for now, we are setting them aside as we try to avoid knowledge-based formulations as much as possible in order to stay within the paradigm of a cost-effective and generic approach in order to reduce deployment costs.

\subsection{Identification procedure}
Let us now successively consider the identification of the process by a linear dynamic model and  a nonlinear NLARX model respectively.
\paragraph{\textbf{Linear state space estimation }} First, the subspace identification method \cite{ljung1998system}  is used to get a representative state space linear model. This identification is carried out through the Matlab toolbox \cite{matlabid} (\textit{ssest} function). This method uses all of the previously mentioned input signals (cf. Fig. \ref{fig:aaa}). The model is as follows:
\begin{equation}
\begin{gathered}
    x[k+1] = A x[k] + B u[k] \\
    y[k]   = C x[k]
\end{gathered}
\end{equation}

$x$ and $u$ are the state and input signals respectively, and $A$, $B$, and $C$ are matrices characterizing the state transition and the observation function. 
To perform prediction, we use the function \textit{forecast}, which performs an ordinary least square method to estimate the initial state with respect to the size of the given past horizon (we chose 20 time steps).  

\paragraph{\textbf{NLARX estimation}} The nonlinear model considered is defined by:
\vspace{-2pt}
\begin{equation}
    y[k+1] = f_{\theta} (y[k], \cdots y[k-n_a], u[k],  \cdots, u[k-n_b])
\end{equation}
\vspace{-2pt}
This model is learned from the data by considering the following loss function for identification, which is well suited for further design of an MPC control:
\begin{equation}
     \frac{1}{N}\sum_{n=0}^{N-1} \frac{1}{P}\sum_{p=0}^{P-1} \left( \hat{y}[t_0(n)+p | t_0(n), \theta ] - y[t_0(n)+p] \right)^2
\end{equation}
With $\hat{y}(t_0(n)+p | t_0(n), \theta )$, the model prediction given the state at $t_0(n)$ and the exogenous data from $t_0(n)+p $ to $t_0(n)$. The function $f_{\theta}$ is the neural network involved in the NLARX formulation.
The prediction is done recursively with the full future prediction approach \cite{MENDOZASERRANO20141301}. We use the solver ADAM \cite{kingma2014method} (an unconstrained non-linear solver) with the Tensorflow library using the BPTT (Back Propagation Through Time) \cite{Werbos1990}.

We trained both linear and non-linear models with the same MIMO formulation and procedure. A hyper-optimization procedure was used using a cross-validation method to reduce bias with this relatively small dataset. The cross-validation is done with data sections instead of regression entries to avoid overlap between samples. The testing dataset is placed at the dataset extremum to avoid any data leak. All the optimized parameters are detailed in TABLE \ref{tab:opti_nlarx_param} and \ref{tab:opti_ss_param}.

\vspace{5pt}

\begin{minipage}[c]{0.49\textwidth}
\centering
\begin{tabular}{|c | c|}
\hline
 Parameters & Values \\
\hline
 N\# layers & \{1,2,3\}  \\
  N\# unit per layer & \{8,16,32,64\}  \\
  learning rate  & [1e-5, 1e-2]  \\
  batch size  & [16,256] \\
  l2 penalization & \{0.0,1e-4,1e-3\} \\
  loss function horizon & [3, 5, 8, 15] \\
\hline
\end{tabular}
\captionof{table}{optimized NLARX model's parameters}
\label{tab:opti_nlarx_param}
\end{minipage}

\begin{minipage}[c]{0.49\textwidth}
\centering
\begin{tabular}{|c|c|}
\hline
 Parameters & Values \\
\hline 
 $n_x$ & [2, 3, 4]  \\
 focus & \{simulation, prediction\}  \\
 subspace horizon & \{48, 24, -1\}  \\
 noise model & \{None, Estimate\} \\
 \hline
\end{tabular}
\captionof{table}{optimized SS model's parameters}
\label{tab:opti_ss_param}
\vspace{-1pt}
\end{minipage}

\subsection{First numerical results}

\begin{figure}
    \centering
    \includegraphics[width=1.0\linewidth]{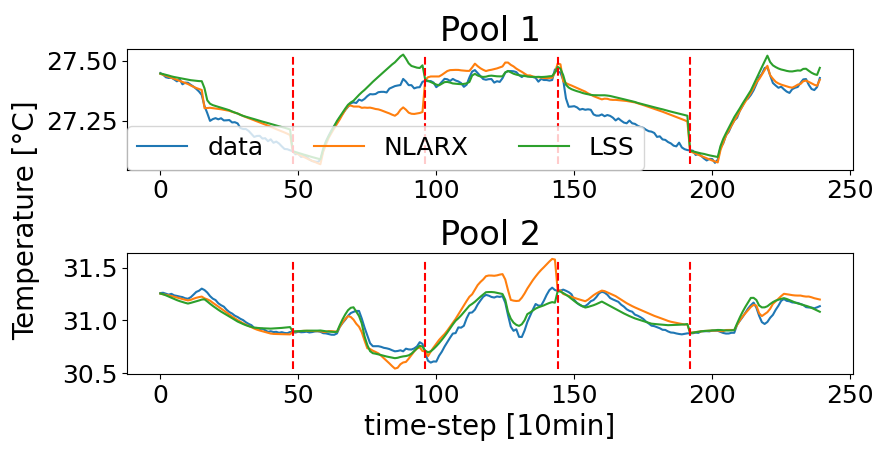}
    \caption{Five 48 steps prediction on test data.}
    \label{fig:test_pred}
       \vspace{-20pt}
\end{figure}

After training, the models are evaluated on both the test and extrapolation sets. In Fig. \ref{fig:test_pred}, we show the temperature predictions for both pools. The dashed red lines depict the time when the predictions for both models are readjusted with the real data. Please note that tools for generating these comparative curves are also available on the benchmark website.

\begin{figure} 
\centering
  \includegraphics[width=0.99\linewidth]{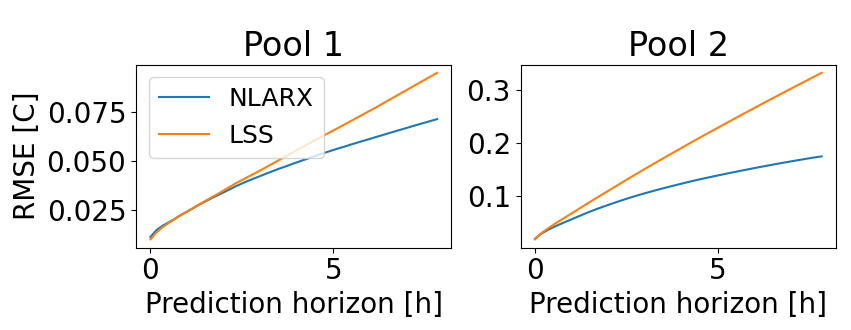}
  \captionof{figure}{RMSE accuracy according to prediction horizon per pool.}
  \label{fig:rmse_model}
     \vspace{-20pt}
\end{figure}

We can now introduce the raw prediction accuracy of the two selected models on the test set (Fig. \ref{fig:rmse_model}). We plotted the accuracy with respect to the prediction horizon and the pool ID. As expected, the accuracy decreases as the prediction horizon increases. The neural network model achieves higher accuracy than the linear one, especially in long-term prediction. This gap is more pronounced for the second pool, which is more challenging to predict due to higher temperatures and more intense activity of people.

Finally, we present the results with the quantitative criteria introduced earlier for both models in TABLE \ref{tab:result_tab}.

\begin{table}[h]
    \centering
           \vspace{15pt}
    \begin{tabular}{|l|c|c|}
         \hline 
         & LSS & NLARX \\
         \hline 
        full horizon acc & 0.32 & \textbf{0.23} \\
        Short term acc & 0.11 & \textbf{0.10} \\
        Long term acc & 0.52  & \textbf{0.31} \\
        Scenario 1 acc & 0.29 & \textbf{0.22} \\
        Scenario 2 acc & \textbf{0.49 }& 0.60 \\
        Scenario 3 acc & \textbf{0.88} & 1.24 \\
        Scenario 4 acc & \textbf{0.36 }& 1.45 \\
        \hline
    \end{tabular}
    \caption{Accuracy score for the proposed models}
    \label{tab:result_tab}
\end{table}

From the table, we can see that the NLARX model has a better accuracy compared to the linear model. However, this difference is less noticeable in short-term accuracy, which highlights the need to consider horizon accuracy for better analysis in predictive control applications. Despite the good results obtained on the test set, a significant decrease in accuracy is observed in all extrapolation scenarios for the non-linear model compared to the linear model. The linear model provides a more consistent level of accuracy between the interpolation and extrapolation regimes.

This first result shows the limits of the pure non-linear model in the extrapolation regime of the learning data. This can be problematic for the control, which will be particularly lacking in resilience in this situation. The linear model seems better adapted to ensure a good regulation of the process considered in this benchmark. However, its accuracy in the training data domain is limited and may induce suboptimal control. This result is expected but underlines the need for a model that combines the qualities of the two models.

\section{Conclusion}

The energy transition requires installations that consume less energy and are more flexible in the face of fluctuations in the energy market. In this context, operating data from a public swimming pool have been made available, to be freely and widely shared. This dataset and evaluation tools have been proposed to serve as a benchmark for learning techniques. 
It should make it possible to design a dynamic black box model, capable of predicting the evolution of the temperature of the pools of the public swimming pool considered. This benchmark will make it possible to compare and, as a corollary, to improve identification tools based on data. It targets patterns for control; with the idea of reducing the cost of deploying advanced control methods.

The characteristics of the target system and the associated datasets are described. The problem was formalized through an input-output representation and evaluation criteria. Matlab-Python functions are offered on the benchmark website. Associated with the proposed datasets, two dynamic models were identified. Their results are quite good in the test set, but they have some weaknesses in the extrapolation regime (especially for the non-linear model). 

In future work, we plan to address the practical limitations of current nonlinear models. More precisely, we aim to develop a method in the spirit of the one presented in \cite{Djeumou2021}. This will involve improving the performance of the nonlinear model by providing it with linear extrapolation capabilities. We encourage the research community to propose the most promising solutions and evaluate them on the benchmark's operating data for comparison purposes.






\bibliographystyle{IEEEtran}
\bibliography{main}
\addtolength{\textheight}{-12cm} 
\end{document}